\documentclass[10pt,twocolumn,letterpaper]{article}

\usepackage{iccv}
\usepackage{times}
\usepackage{epsfig}
\usepackage{graphicx}
\usepackage{amsmath}
\usepackage{amssymb}
\usepackage{lipsum}

\let\svthefootnote\thefootnote

\newcommand\blankfootnote[1]{%
  \let\thefootnote\relax\footnotetext{#1}%
  \let\thefootnote\svthefootnote%
}
\let\svfootnote\footnote
\renewcommand\footnote[2][?]{%
  \if\relax#1\relax%
    \blankfootnote{#2}%
  \else%
    \if?#1\svfootnote{#2}\else\svfootnote[#1]{#2}\fi%
  \fi
}

\iccvfinalcopy 

\begin{document}

\title{DynamoNet: Dynamic Action and Motion Network}

\author{
    {  Ali Diba$^{1,\star}$,  Vivek Sharma$^{2,\star}$,  Luc Van Gool$^{1,3}$, Rainer Stiefelhagen$^{2}$}\\
    {\normalsize {$^{1}$ESAT-PSI, KU Leuven, $^{2}$CV:HCI, KIT,  $^{3}$CVL, ETH Z\"{u}rich}} \\ 
     \tt\small  \{firstname.lastname\}@esat.kuleuven.be, \{firstname.lastname\}@kit.edu  
}

\maketitle

\begin{abstract}
In this paper, we are interested in self-supervised learning the motion cues in videos using dynamic motion filters for a better motion representation to finally boost human action recognition in particular. Thus far, the vision community has focused on spatio-temporal approaches using standard filters, rather we here propose dynamic filters that adaptively learn the video-specific internal motion representation by predicting the short-term future frames.  We name this new motion representation, as dynamic motion representation (DMR) and is embedded inside of 3D convolutional network as a new layer, which captures the visual appearance and motion dynamics throughout entire video clip via end-to-end network learning. Simultaneously, we utilize these motion representation to enrich video classification. We have designed the frame prediction task as an auxiliary task to empower the classification problem.

With these overall objectives, to this end, we introduce a novel unified spatio-temporal 3D-CNN architecture (DynamoNet) that jointly optimizes the video classification and learning motion representation by predicting future frames as a multi-task learning problem. We conduct experiments on challenging human action datasets: Kinetics 400, UCF101, HMDB51. The experiments using the proposed DynamoNet show promising results on all the datasets.
\end{abstract}

\section{Introduction}
\footnote[]{$^{\star}$Ali Diba and Vivek Sharma contributed equally to this work.
}

Human action recognition~\cite{i3d,stcnn,t3d,tle,idt} in videos has attracted a huge attention over the last decade, due to the potential applications in video surveillance, understanding, analysis, retrieval tasks and more. In practice, the performance of computer vision systems still falls behind that of humans. On top of the challenges that make object class recognition hard, complicating aspects like camera motion and the continuously changing viewpoints negatively influences the vision system. Although,  Convolutional Neural Networks (ConvNets) have successfully upsurged several sub-fields of vision, for action recognition particularly, they still lack to model the motion cues effectively.

 \begin{figure}[t]
\centering
{\includegraphics[width=1\columnwidth]{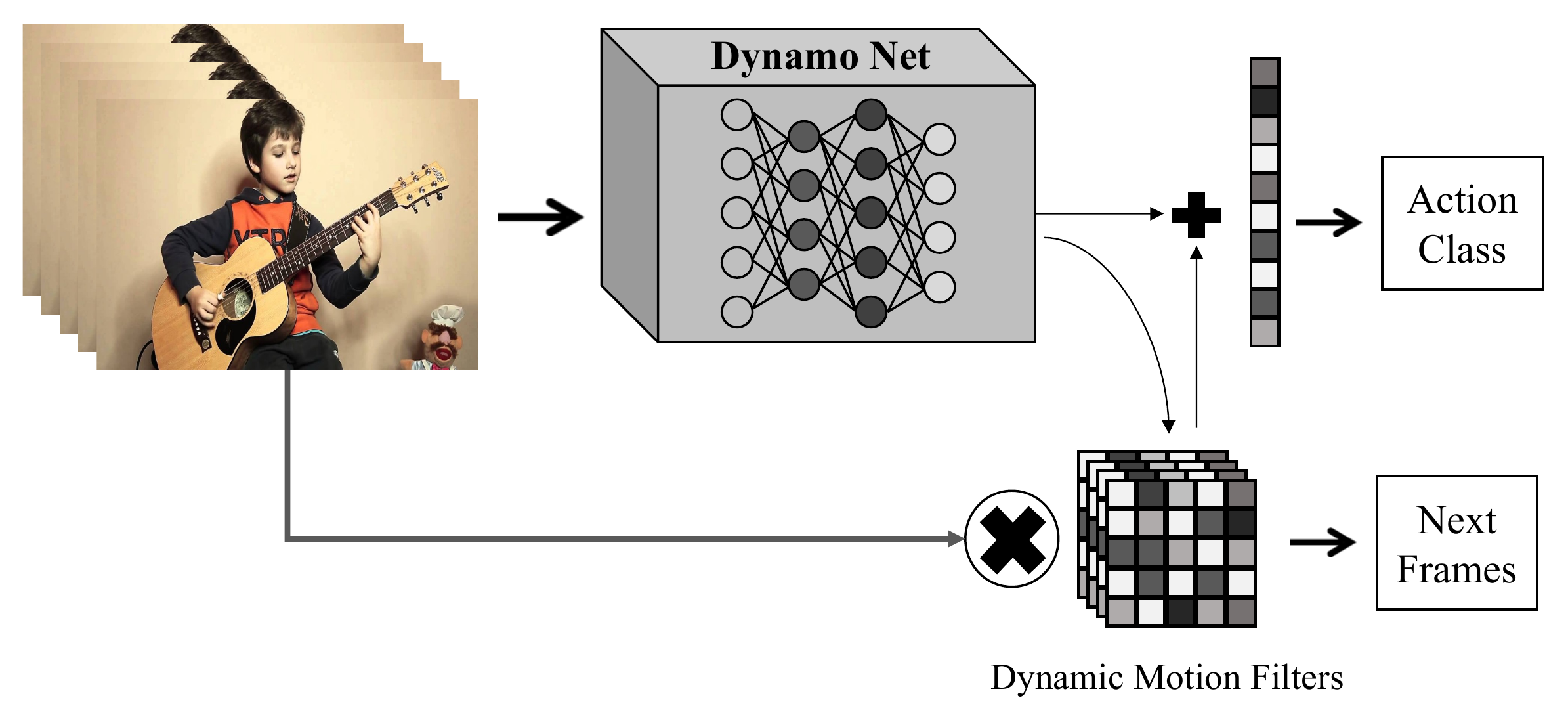} 
} 
\caption{Overview of the proposed multi-task learning ConvNet architecture to simultaneously classify action and learn motion representation by predicting future frames in end-to-end learning manner.  $\times$, $\mathbf{+}$ denotes convolution and concatenation operations.} \label{fig:front}
\vspace{-0.45cm}
\end{figure}

Neural networks for action recognition can be mainly categorized into two types, namely \textit{architecture-driven} ConvNets~\cite{twostream,c3d} (which utilize standard filters and pooling kernels in a video architecture to exploit long-range dynamics) and~\textit{encoding-driven} ConvNets~\cite{tle,actionvlad} (which integrate new encoding methods in addition to standard filters and pooling kernels in a video  architecture to learn spatio-temporal feature representation). This paper falls in the pool of~\textit{encoding-driven} ConvNets, where we extend the standard video understanding architecture to incorporate a new layer to learn dynamic motion representation conditioned on the~\textit{action-specific} basis.  

In this paper, we propose to extend the training of ConvNets-based action classification to incorporate the high-level goal to learn \textit{action-specific} motion representation via future frame prediction. Our contribution named as DynamoNet is a method that jointly optimizes a ConvNet for video action classification and future frame prediction as a multi-task learning problem. We achieve this by adaptively learning the motion features on a \textit{video-specific} basis via dynamic motion filters, which enables the motion prediction model to selectively utilize only those motion features that lead to improved video classification. 


Since we understand the vital role of motion feature representation, we propose to use the dynamic convolutional filters to dynamically 
discover and learn \textit{video-specific} internal motion representations for improved video classification (see Fig.\ref{fig:front} for a graphical overview). Our paper is inspired from~\cite{dfn,dyncnn}. However, while Brabandere et al.~\cite{dfn}  applies the dynamic filters to transform an angle to a filter (steerable filter) and Sharma et al.~\cite{dyncnn}  emulates a range of enhancement filters to generate image enhancement methods. We used the same terminology as in~\cite{dfn,dyncnn}. In contrast to these works, our work differs substantially in technical approach and the application scope.
Our DynamoNet is designed to learn  motion representation, which we achieve by adaptively extracting informative features by predicting the short-term future frames  to improve classification.
We believe predicting (or reconstructing) the future frames selectively transfers the \textit{fundamental notion} of motion content to the filters, which in turn improve the overall effectiveness of the motion representations. The motion representation learning is jointly optimized along with the video classification as a multi-task learning problem. Using short-term future frame prediction as a proxy task is promising, and we clearly show that this works for an accurate action recognition in videos.

Precisely, our network takes the current video-clip with $T$ frames and generates $T+1$ future frames using  $T$ dynamic motion kernels or dynamic filters. The network structure is based on 3D ConvNets. Specifically, given an input video-clip $x$ with $T$ frames the network generates $T$ dynamic motion kernels ($F_{T}$) to predict the consecutive next frame given the previous one i.e. $F_{t}: x_{t} \rightarrow \hat x_{t+1}$, $t \in \{1,\ldots,T\}$. These filters are video dependent motion kernels and are conditioned on the input and therefore vary from one sample to another during training and testing phase, which means that the filter dynamically extracts important motion representation from a given input. Further, we utilize these $T$ $d$-dimensional dynamic motion features along with STC-ResNext~\cite{stcnn,resnet3d} 3D-ConvNet features for video classification. Moreover, we believe the dynamic motion kernels capture the important concepts of motion representation from the temporal cue and the extracted motion features are robust and compact global temporal representation for the whole video, this makes them a perfect fit for action recognition task well. 
Our method is evaluated on three challenging benchmark action recognition datasets, namely UCF101, HMDB51 and Kinetics 400. We experimentally show that the 3D ConvNets when combined with our dynamic motion filters (see Sec.~\ref{sec:exp}) achieve state-of-the-art performance on UCF101 (97.8\%), HMDB51 (76.8\%) and Kinetics 400 (77.9\%).


\section{Background and Related Work} \label{sec:related}
\noindent
\textbf{Action Recognition with ConvNets.} 
With Convolutional Neural Networks (ConvNets) the vision community has successfully made huge leaps forward in several sub-fields of vision and has outperformed hand-engineered representations by a significant margin particularly for action recognition. End-to-end ConvNets have been introduced in~\cite{pooling,karpathy,twostream,c3d,tsn}  to exploit the appearance and the temporal information. These methods operate of 2D (individual image-level) or 3D (video-clips or snippets of $K$ frames).  In the 2D setting, spatial and/or temporal information are modeled via LSTMs/RNNs to capture long-term motion cues~\cite{lstm1,n3d}, or via feature pooling and encoding methods  using Bilinear models~\cite{tle}, Vector of Locally Aggregated Descriptors~(VLAD)~\cite{actionvlad}, and Fisher vector encoding~(FVs)~\cite{fishernet}. While, in the 3D settings, the input to the network consists of either RGB video clips or stacked optical-flow frames to capture the long-term temporal information. The filters and pooling kernels for these architectures are 3D (x, y, time) i.e. 3D convolutions~($s \times s\times d$)~\cite{n3d} where $d$ is the kernel temporal depth and $s$ is the kernel spatial size. Simonyan et al.~\cite{twostream}  use two-stream 2D ConvNets cohorts of RGB images and a stack of 10 optical-flow frames as input. Tran et al.~\cite{c3d} on the other side explored 3D ConvNets with fixed kernel size of  $3\times 3 \times 3$, where spatio-temporal feature learning for clips of 16 RGB frames was performed. In this way, they avoid to calculate the optical flow explicitly and still achieve good performance. Further, in~\cite{res3d,r2plus1} Tran et al. extended the ResNet architecture with 3D convolutions. In~\cite{pooling} Feichtenhofer et al. propose 3D pooling.  Sun et al.~\cite{sun3d} decomposed the 3D convolutions into 2D spatial and 1D temporal convolutions. Wang et al.~\cite{tsn} propose to use sparsely sampled non-overlapping frames from the whole clip as input for both spatial and temporal streams and then combine their scores in a late fusion approach.  Carreira et al.~\cite{i3d} propose converting a pre-trained 2D ConvNet~\cite{googlenet} to 3D ConvNet by inflating the filters and pooling kernels with an additional temporal dimension $d$. All these architectures have fixed temporal 3D convolution kernel depths throughout the whole architecture. In T3D~\cite{t3d}, Diba et al. propose temporal transition layer that models variable temporal convolution kernel depths over shorter and longer temporal ranges. Furthermore in~\cite{stcnn}, Diba et al. propose spatio-temporal channel correlation that models correlations between channels of a 3D ConvNets wrt. both spatial and temporal dimensions. In contrast to these prior works, our work differs substantially in scope and technical approach. We propose an architecture to learn dynamic motion filters for modeling an effective internal motion representation in order to adaptively extract informative motion features conditioned on a video-specific basis to improve action recognition.

Finally, it is worth noting the self-supervised learning works on ``harvesting" training data from unlabeled sources for action recognition.  Fernando~\etal~\cite{Odd-one} and Mishra~\etal~\cite{Shuffle} shuffle the video frames and treat them as positive/negative training data; Sharma et al.~\cite{sharmafg} mines labels 
using a distance matrix based on similarity although for video face clustering; Wei et al.~\cite{wei} divides a single clip into non-overlapping 10-frame chunks, and then predict the ordering task; Ng et al.~\cite{actionflow} estimates optical flow while recognizing actions. We compare all these methods against our unsupervised future frame prediction based ConvNet training in the experimental section.



\noindent
\textbf{Future Frame Prediction.} Given an observed image or a sequence of frames predict the future frames is very popular these days~\cite{wu34,wu21,wu33,shi2018action,wu32,wu23,wu22,wu24}, where mostly researchers predict low-level pixels or motion, until recently image synthesis is done via neural networks: Generative adversarial networks~\cite{wu26,wu25,wu6,wu35,vondrick2017generating}, variational auto-encoders~\cite{wu7,xue2018visual,wu8},  deep regression networks~\cite{wu19} to anticipate the visual representation in the future. These works basically forms to be a part of either deterministic prediction frameworks~\cite{wu11,wu19} or probabilistic prediction frameworks~\cite{wu35,xue2018visual}.  It has been shown that the probabilistic content-aware motion prediction models the motion fields or image features better in comparison to deterministic models which cannot model the uncertainty well. Successful predicting the future will demonstrate the computers understanding of objects in the scene.  One straight forward approach would be discriminate training to predict the next future frames, but our world is full of uncertainty. Its not the predictions are wrong, its just an intrinsic ambiguity in the prediction. In contrast to previous work, our main goal is not to predict future frame, but rather to learn motion representations by predicting future frames. Precisely, we propose a method that predicts high-level concepts such as objects and actions by learning dynamic motion filters to predict the consecutive next frame given the previous one in a self-supervised manner.

Further, in the similar spirit of frame prediction although for different tasks using action recognition datasets, it is worth noting works of Mathieu et al.~\cite{mathieu2015deep} and Srivastava et al.~\cite{srivastava2015unsupervised}. We quantitatively compare to Mathieu et al.~\cite{mathieu2015deep} in our experiments. 

\noindent
\textbf{Filter Generating Networks.} Starting with the seminal work of Jaderberg et al.~\cite{stn} where the authors propose transformation filters to do translation and rotation, all these papers~\cite{dfn,dcl,dyncnn} utilize the same concept (deep-down) to learn a steerable filter~\cite{dfn}, weather prediction filter~\cite{dcl}, or an enhancement filter~\cite{dyncnn} using input-output image pairs. Different from these works, we propose to apply these filters for learning motion representation in videos with an overall goal to improve action classification. We clearly show that this works in our experiments.

\section{Proposed Method}  \label{sec:app}
Our aim is to learn a dynamic motion representation model with an overall goal to improve video classification. To this end, we propose two ConvNet architectures described in this section.
Our first architecture is proposed to learn  dynamic motion filters by predicting the short-term future frames in an end-to-end self-supervised learning fashion. The other proposed end-to-end network is designed to simultaneously classify action, in addition to learn motion representation by predicting the future frames. 

We use 3D ConvNets~\cite{resnet3d,stcnn}, STCnet/3D-ResNext architecture as the base models, and incorporated two branches to do classification and frames prediction. The input to the network is a stack of 16, 32 or 64 frames in different experimental setups, which we refer as a video clip.

\subsection{Dynamic Motion Filters} \label{subsec:dmf}
The filter generating network (DynamoNet) is inspired from~\cite{dfn,dcl,dyncnn} and is composed of 3D filters and pooling kernels, with the last fully-connected layer (i.e., dynamic motion filter parameters). The DynamoNet is self-supervised or unsupervised. The DynamoNet maps the input to the filter. Precisely, the network takes a video-clip $x$ with $T$ frames and outputs filters $F_{\Theta,t}, \Theta \in \mathbb{R}^{s \times s \times t}$, where $\Theta$ is the transformation parameter that learns the mapping, $s$ is the spatial kernel size and $t$ is the number of filters - which is driven by the input stack of frames i.e. $t \in T$. 

\begin{figure*}[ht]
 \centering
 \includegraphics[width=1.99\columnwidth]{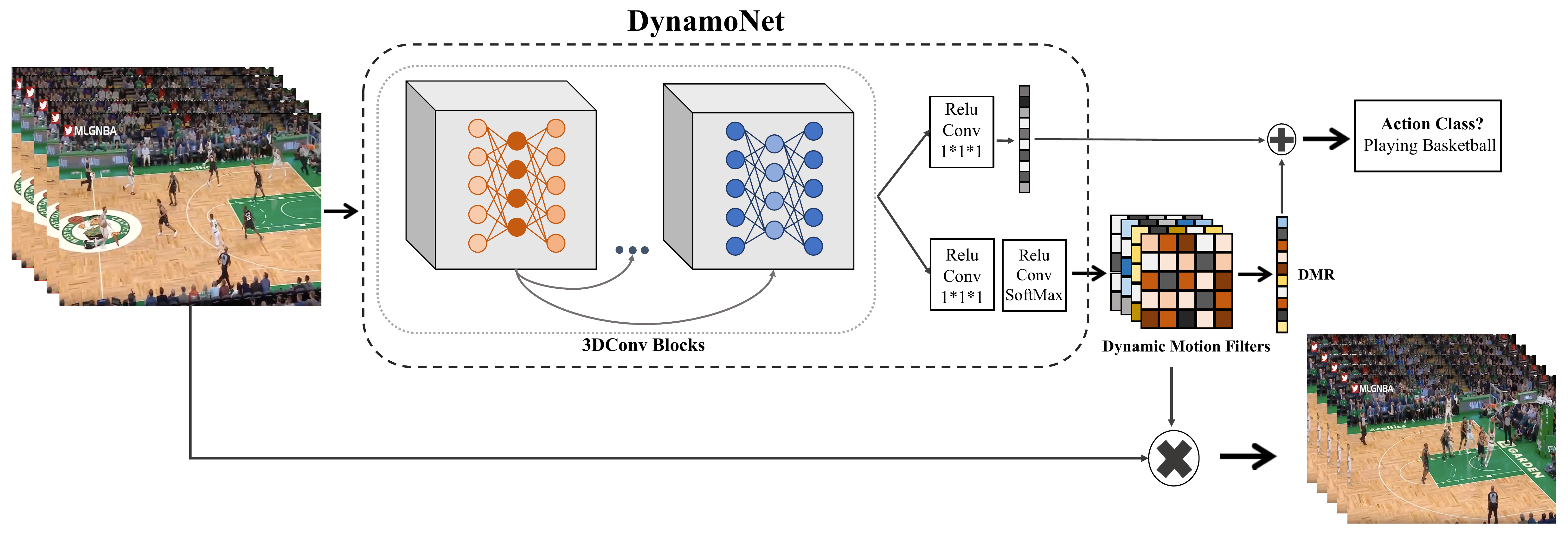}
 \caption{\textbf{DynamoNet.} Input to the network is a video with $T$ frames, that generates $T$ dynamic motion filters to predict the consecutive next frame given the previous one to learn motion representation. The motion filters are then concatenated together to form a global representation, along with the STC-ResNext features and then fed to classifier. The network is jointly optimized with a classification objective to adaptively extract informative motion features for improved classification. $\times$, $\mathbf{+}$ denotes convolution and concatenation operations.} 
  \label{fig:dynamonet}
  \vspace{-0.45cm}
 \end{figure*}

Given an input video-clip $x \in \mathbb{R}^{H\times W \times T}$, the network produces dynamic motion filters to predict the consecutive future frames $\hat x \in \mathbb{R}^{H\times W \times (T+1)}$ where $H$, $W$ denotes the frame height and width. The scheme is illustrated in Fig.~\ref{fig:dynamonet}. The future frame predictor network can be formulated as:
\begin{equation}
\hat x_{t+1} = F_{\Theta,t}(x_{t})
\end{equation}

The $F_{\Theta,t}$ motion filters are convolved with the input $t^{th}$ frame $x_{t}$  to generate $\hat x_{t+1}$ frame. $F_{\Theta,t}$ is applied to $x_{t}$  at every spatial position ($H,W$) to output the predicted frame $\hat x_{t+1} \in \mathbb{R}^{H\times W}$. Note that the filters are sample-specific and are conditioned on the input $x_{t}$. In Fig.~\ref{fig:dynamonet}, we show filter generating network to predict dynamic filters and thus predicting the future frames. The filter size determines the receptive field and is application dependent. From the literature~\cite{dyncnn}, we exploited a lot of insights about kernel sizes. 
For motion prediction, we have tested with different filter sizes $s = \{3,4,5,6,7\}$ and for our setup we found that a filter size of $5\times5$ gave the best result, and others ($> 5\times5$ or  $< 5\times5$) produced smoother images, with a drop in classification performance by approx $\sim$2-3\%. Further, we found that frame prediction done at deep-level performed better in comparison to the intermediate levels.

For generating the motion filter parameters, the network is trained using Huber loss function~\cite{huberloss} between the target future frame $x_{t+1}$ and the network's predicted future frame $\hat x_{t+1}$. The frame prediction~(FP) using the Huber loss function is defined as:
\begin{equation}
\mathcal{L}_{FP} =\left\{
\begin{array}{ll}
\frac{1}{2}||\hat x_{t+1}-x_{t+1}||^{2}_{2}\quad \textbf{if} \quad
||\hat x_{t+1}-x_{t+1}||_{1} < \delta \\
\\
\delta||\hat x_{t+1}-x_{t+1}||_{1} - \frac{1}{2}\delta^{2}\quad otherwise
\end{array}
\right.
\end{equation}

where the threshold $\delta$ is set to $0.01$. More details on the training and the network architecture is discussed in the experimental section.

Our future frames prediction method differs from all the recent methods, as  we utilize motion filters to synthesize the next frames given the previous one, $F_{t}: x_{t} \rightarrow \hat x_{t+1}, t \in \{1,\dots,T\}$. The main difference of our method for generating next frames in comparison to other methods differs substantially in technical approach, we do not generate frames directly by conv-deconv layers, rather we use these layers to generate motion filters to reconstruct and predict the next frames. We believe predicting (or reconstructing) the future frames selectively transfers the \textit{fundamental notion} of motion content to the filters, which in turn improve the overall effectiveness of the motion representations. Furthermore, in this way, our network configuration and the learning scheme helps to learn the pixel motion information in an spatio-temporal regime.


Inspired by dynamic filter networks~\cite{dfn}, we believe for learning an effective motion representation from a video, dynamically generated filters is a robust solution to  discover and capture video-specific internal motion variations. As the parameters of the filters are conditioned on the input, they vary from one sample to another which is perfect for learning internal motion representations and variations in the video. By extracting this intra-information from the clip, we believe we model motion representation from the same clip, since the generated filters are demonstrating the dynamic information of frames. This representation can be effective for video classification task, in the next section we show how joint optimization can be modeled for motion representation learning and classification - simultaneously to have a more solid action classifier.

\noindent
\textbf{Example frame prediction structure}: 
Let's assume that a given input clip has 16 frames, the predicted future frames are 16 frames consisting 15 reconstructed ones (2nd to 16th frames from input) and a new future frame obtained from the 16th frame. Each predicted (reconstructed) frame is obtained by applying the dynamic motion filter to the previous frame (input frame) at every spatial position. In this way, the dynamic motion filters are learned by predicting the consecutive next frame given the previous one.

\subsection{Action Recognition}
We utilize 3D ConvNets over 2D ConvNets for video classification and frame prediction because of their compelling advantages of exploiting long-range temporal cues rather than merely spatial cues. Precisely, we use the recently proposed STCnet~\cite{stcnn} or 3D-ResNet/ResNext~\cite{resnet3d} as the main building block of the DynamoNet. We chose these architectures because of their promising performance for action classification both in terms of accuracy and high computation speed. Further, the 3D temporal convolution kernels efficiently capture the visual appearance and the temporal information  across frames in videos in comparison to 2D convolutions - which lacks to model the temporal dimension. That being said, 3D-STCnet/ResNet are good candidates to extract the spatio-temporal feature representations for action classification and motion analysis.

Here, we recycle the dynamic motion filters architecture from Section~\ref{subsec:dmf}. Figure~\ref{fig:dynamonet} shows the schematic layout of the whole architecture. Our architecture has two network branches, one branch learns the dynamic motion representations $F_{\Theta,T}$ and the second branch is the standard fully-connected layer of the 3D-STCnet~(AR). Both of the network branches are trained together, therefore we have the action representation~(AR) and dynamic motion representation learned at the same time.
As the last step to action classification, we flatten the motion filters and then followed by a fully-connected layer of size $\mathbb{R}^{d}$, the $d$-dimensional DMR is then concatenated with AR and then fed to classification layer. By this design we incorporate the motion information for action classification task.

\textbf{End-to-end learning.} Finally, we now extend the loss of approach~1, by adding the softmax-loss (classification)  for joint optimization of motion filter learning by future frame prediction with a classification objective. The total loss of the whole pipeline is given by:
\begin{equation}
\mathcal{L}_{total} =\alpha\mathcal{L}_{FP} + \beta\mathcal{L}_{Classification} \\
\end{equation}

where $\alpha$, $\beta$ are losses weights and both of the tasks are optimized together. We show it both qualitatively and quantitatively that the trained DMR is optimized to capture an accurate motion information for sample-specific actions in this manner. Figure~\ref{fig:qresult} shows qualitative results with dynamic filter based predicted frames.

\subsection{Unsupervised Training}
For comparison with the previous self-supervised or unsupervised representation learning methods~\cite{Odd-one,Shuffle,actionflow,wei}, we remove the classification branch and just keep the dynamic motion filters with the frame prediction part only, an unsupervised video learning pipeline is obtained. As we already know, the important aspect of videos are meaningful motions. Using our approach, in practice, one can easily learn motion representation in an unsupervised way by simple reconstructing and/or predicting the future frames via the self supervisory signal available in the sequence of frames.  

We have investigated the method with a number of unlabeled videos and trained the network from scratch. We show that such a unsupervised pre-training can be very beneficial for a stable model weight initialization, and  thus this reduces the need of large labeled video datasets for training 3D ConvNets from scratch for the action classification task. More details on the training schemes and their results are discussed in the experimental section.

\section{Experiments} \label{sec:exp}

In this section, we first introduce the datasets, implementation details of our proposed approach, and then show the applicability of unsupervised pre-training, followed by the role of frame prediction in the training scheme. Finally, we compare our method with the state-of-the-art methods on three challenging human action and activities datasets.

\subsection{Network Design}
The DynamoNet consists of three parts: first is the 3D-Conv which in our experiments is STCnet or 3D-ResNet/ResNext with different depths. We have applied more layers for two branches, for the action classification part to extract an efficient representation, we added two layers of convolution and a fully connected linear layer. On the frame prediction, there are two Conv layers with a softmax layer to yield the dynamic filters. After flattening the filters, there is a fully connected layer with size of 512 to extract the dynamic motion representation. The AR and DMR features are concatenated together and then fed to the classification loss. In the rest of this section, for the DynamoNet~(STCnet) we use STC-ResNext101 and for DynamoNet~(ResNext) we use ResNext101.

\subsection{Datasets}
We evaluated our proposed DynamoNet on three challenging human action and activities datasets; HMDB51~\cite{hmdb51}, UCF101~\cite{ucf101} and Kinetics~\cite{i3d}. We use the pre-defined training/testing splits and protocols provided originally. We report the mean average accuracy over the three splits for HMDB51 and UCF101 and for Kinetics, we report the performance on the validation set.

\textbf{Kinetics.}
Kinetics is a challenging human action recognition dataset introduced by~\cite{i3d}, which contains 400 and 600 action classes. There are two versions of this dataset: untrimmed and trimmed. The untrimmed videos contain the whole video in which the activity is included in a short period of it. However, the trimmed videos contain the activity part only. We evaluate our models on the trimmed version. We use all training videos for training our models from scratch.

\textbf{UCF101.}
To evaluate our DynamoNet action recognition performance, we first trained it on the Kinetics dataset and then fine-tuned  on UCF101. Furthermore, we also evaluate our models by training them from scratch on UCF101 using randomly initialized weights and unsupervised pre-training method to study the impact of pre-training on a huge dataset such as Kinetics, and also the unsupervised pre-training method. 

\textbf{HMDB51.}
For HMDB51, we employ the same methodology as UCF101 and we fine-tune the DynamoNet on HMDB51, which was pre-trained on Kinetics. Also, we similarly evaluate our model by training it from scratch on HMDB51 using randomly initialized weights.

\subsection{Implementation Details}
We have used the PyTorch framework for the implementation and all the networks are trained on 8 P100 NVIDIA GPUs. Here, we describe the implementation details of our proposed DynamoNet, frame prediction and action classification.

\textbf{Training.} We train our DynamoNet with just frame prediction part as the pre-training step on 500K unlabeled video clips from YouTube8M dataset. DynamoNet operates on a stack of 16 or 32 RGB frames. We have resized the video frames to 122px when smaller and then randomly do the 5 crops (and their horizontal flips) of size $112\times112$ as the main network input size.  For the network weight initialization, we adopt the same technique proposed in~\cite{densenetweighting}. For the model training, we use SGD, Nesterov momentum of 0.9, weight decay $10^{-4}$ and batch size of 64.  The learning rate for start is set to 0.1 and reduced by a factor of 10 manually when the validation loss is saturated.  To train the total loss, we set the coefficients of the losses as: $\alpha=0.1$ and $\beta=1.0$. Once the unsupervised pre-training is done, the main training with both branches of action recognition and frame prediction is done on Kinetics dataset with a maximum number of 200 epochs. We also employ batch normalization for network training. In our experiments, we use different version of STCnet and 3D-ResNet/ResNext as the main convolutional section of DynamoNet, since they are state-of-the-art methods in 3D-CNN action models. We have evaluated different depth of these networks in our experiments. STCnet has similar structure of 3D-ResNet with an extra module to handle spatio-temporal channel correlations in Conv layers.

\textbf{Testing.}  For action recognition on videos, we separate each video into non-overlapping clips of 16/32/64 frames. The DynamoNet is applied over the video clips by taking a $112\times112$ center-crop, and for a video-level prediction, we average the prediction scores over all clips in a video.

\subsection{Unsupervised Pre-Training}
Since the frame prediction part can be trained separately without the need of labeled video, we have studied the effect of unsupervised pre-training the 3D-Conv part of network which carries most of information. As mentioned before the network is trained for frame prediction on 500K unlabeled video clips from YouTube8m. While doing pre-training, the action classification part is detached. After the pre-training is done, both of the branches are activated to be trained also for action classification as well. 


We fine-tune with lower learning rate the self-supervised (or unsupervised) pre-trained network on UCF101/HMDB51 directly, and in Table~\ref{table:ucfS} we show that our method performs better in comparison to state-of-the-art self-supervised methods~\cite{Odd-one,Shuffle,actionflow,wei} when trained from scratch on UCF101/HMDB51. It is obvious that our DynamoNet used more data to train, but the extra data is just video clips without any labels. So our method is an effective way to do pre-training without any cost on data labeling for video classification methods.

\begin{table}[t]

{\small
\tabcolsep=0.3cm
\begin{center}
\begin{tabular}{ l |   c|c}
\hline
Model&   UCF101& HMDB51\\
\hline
\hline
3D-ResNet 101     & 55.4& {29.2}\\
STC-ResNet 101      & {56.7}& {30.8}\\
\hline
Shuffle and learn~\cite{Shuffle}      & {50.9} & {19.8} \\
Odd-One-Out~\cite{Odd-one}      & {60.3} & {32.5}\\
ActionFlowNet~\cite{actionflow}   & {83.9} & {56.4} \\
AOT~\cite{wei}      & {86.5} & {-} \\
\hline
DynamoNet~(ResNext)     & \textbf{87.3}& \textbf{58.6}\\
DynamoNet~(STCnet)      & \textbf{88.1}& \textbf{59.9}\\
\hline
\end{tabular}
\end{center}}
\caption{Comparison of self-supervised methods on UCF101 and HMDB51 split-1 with RGB input. All of the methods (except baseline networks) has been trained with a self-supervised method and then fine-tuned on UCF101 and HMDB51.}
\label{table:ucfS}
\vspace{-0.25cm}
\end{table}

\begin{figure*}[ht]
 \centering
 \includegraphics[width=1.99\columnwidth]{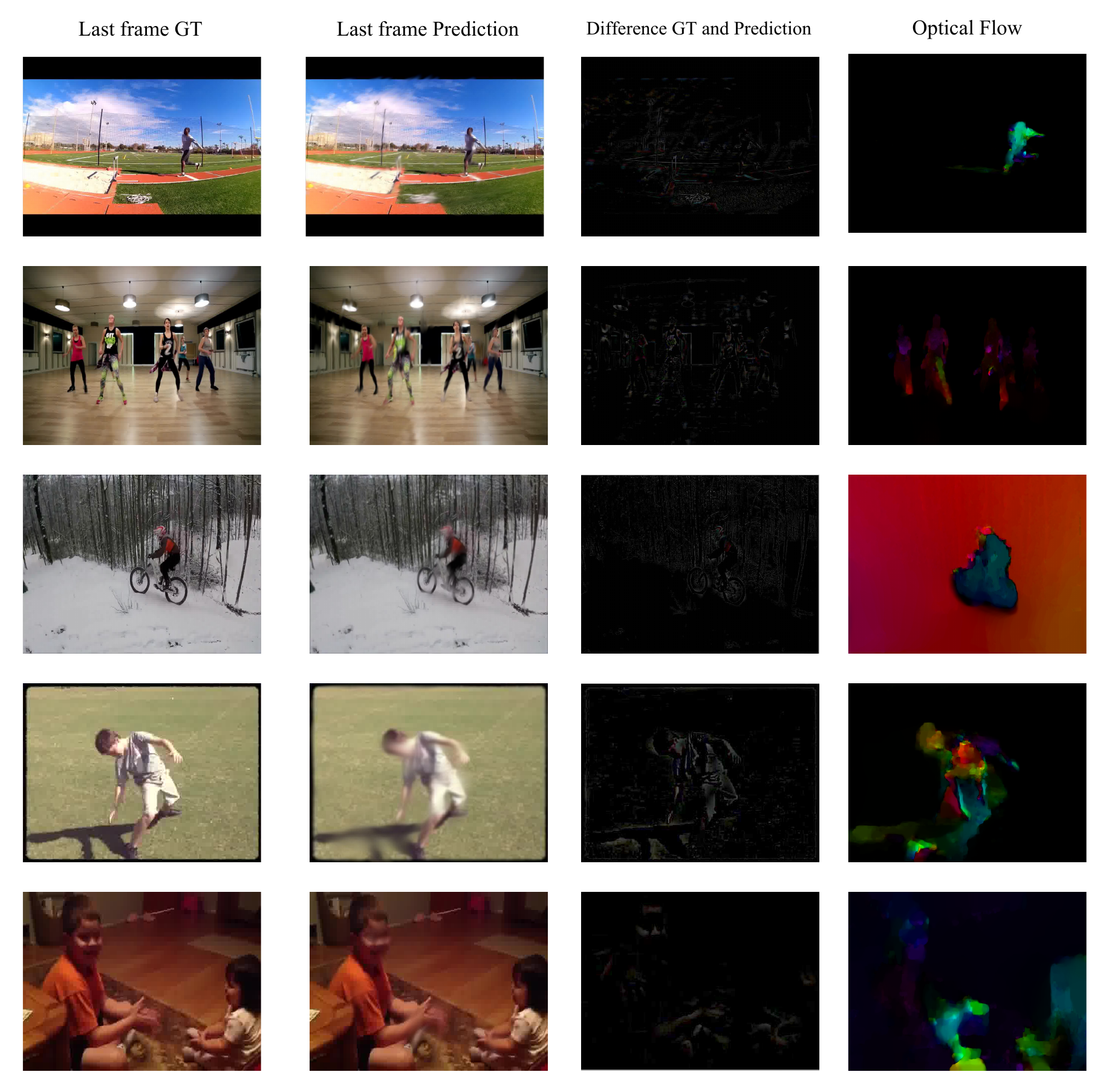}
 \caption{\textbf{Qualitative results.} Comparison between the actual ground truth frame and the predicted future frame - obtained using our proposed approach. First column is an actual frame from video clip as Ground-Truth~(GT). Second column is the DynamoNet predicted frames. Third is the difference image of the GT and the predicted frames, and in fourth column, we show the optical flow extracted from the predicted frames which presents corresponding motion regarding the last frame. Best viewed in color.} 
  \label{fig:qresult}
 \vspace{-0.45cm}
 \end{figure*}

We have also evaluated the impact of the self-supervised (or unsupervised) pre-training on the demand of labeled training data for training on large datasets like Kinetics. To train a 3D-CNN like STCnet or 3D-ResNet from scratch, we need a huge amount of labeled video clips. We showed with a pre-trained 3D-ConvNet by DynamoNet pipeline, we can use a fraction of Kinetics videos but still achieve a reasonable performance. Table~\ref{table:kinUN} shows how we can handle the cases with limited number of videos. The evaluation is done on the validation set of Kinetics. We can observe that DynamoNet performance  when trained with half of the dataset is still comparable with others trained on full amount of data.

\begin{table}[hbt]

{\small
\tabcolsep=0.5cm
\begin{center}
\begin{tabular}{ c |c  |c}
\hline
Model & Data size & Top1-Val~(\%)\\
\hline
\hline
3D-ResNext 101 \cite{3DResHara}& half     & 53.9\\
3D-ResNext 101 \cite{3DResHara}& full     & 65.1\\
STC-ResNext 101  \cite{stcnn}& half     & {55.4}\\
STC-ResNext 101  \cite{stcnn}& full     & {66.2}\\
\hline
DynamoNet~(STCnet) & half   & \textbf{63.6}\\
DynamoNet~(STCnet) & full     & \textbf{67.6}\\
\hline
\end{tabular}
\end{center}}
\caption{Evaluation of training on half and full Kinetics dataset. }
\label{table:kinUN}
\vspace{-0.25cm}
\end{table}

\begin{table}[hbt]
{\small
\tabcolsep=0.5cm
\begin{center}
\begin{tabular}{ l |c}
\hline
Model &  SSIM~\\
\hline
\hline
Mathieu et al.~\cite{lecUcf101}&  0.92 \\
Ours&  \textbf{0.95}\\
\hline
\end{tabular}
\end{center}}
\caption{Frame prediction quantitative performance comparison on 378 test videos from UCF101.}
\label{table:pred_result}
\end{table}

\subsection{Frame Prediction Learning Impact}
There might be a question about the effect of prediction loss on the training pipeline. We have done experiments to generate filters and have used them as feature combined with action representation for action classification without frame prediction objective. A DynamoNet which is trained on the UCF101 achieves 50.2\%, training DynamoNet without prediction loss performs 43.2\%. As expected, the performance was poor and filters do not present any meaningful information in absence of prediction objective function.


\subsection{Frame Prediction}
We have compared our frame prediction performance with Mathieu et al.~\cite{lecUcf101} which provided results on 378 test videos from UCF101 in Table~\ref{table:pred_result}. Moreover in Figure~\ref{fig:qresult}, we present a few examples of qualitative prediction results using our DynamoNet.
The predicted frames show that the dynamic motion filters are able to capture the motion information, and thereby help to predict the future frames.

\subsection{Action Recognition}

\begin{table}[htb] 

{\small
\tabcolsep=0.2cm
\begin{center}
\begin{tabular}{ l |   c |  c }
\hline
\textbf{Method} &  \textbf{Top1-Val} &  \textbf{Top5-Val} \\
\hline
DenseNet3D            & 59.5 & -\\ \hline
Inception3D             & 58.9 &-\\ \hline
C3D~\cite{3DResHara}      & 55.6  &-\\ \hline
3D ResNet101~\cite{3DResHara}   & 62.8  & 83.9\\ \hline
3D ResNext101~\cite{3DResHara}    & 65.1  & 85.7\\ \hline
RGB-I3D~\cite{i3d}    & 68.4  & 88\\ \hline

{STC-ResNet101 (16 frames)} \cite{stcnn}        & 64.1  &85.2\\ \hline
{STC-ResNext101 (16 frames)} \cite{stcnn}         & 66.2  &86.5\\ \hline
{STC-ResNext101 (32 frames)} \cite{stcnn}         & {68.7}  & {88.5}\\ 
\hline\hline
{DynamoNet~(ResNext) (16 frames)}         & {66.3}  & {86.7}\\ \hline
{DynamoNet~(ResNext) (32 frames)}         & {68.2}  & {88.1}\\ \hline
{DynamoNet~(STCnet) (16 frames)}          & {67.6}  & {87.2}\\ \hline
{DynamoNet~(STCnet) (32 frames)}          & {69.9}  & {90}\\ \hline
\textbf{DynamoNet~(STCnet) (64 frames)}         & \textbf{77.9}   & \textbf{94.2}\\ \hline
\end{tabular}
\end{center}}
\caption{Performance~(\%) comparison of \textbf{DynamoNet} with other state-of-the-art methods on Kinetics-400 dataset.}
\label{table:kinetics_results}
\end{table}

In this section, we compare the DynamoNet performance with the state-of-the-art methods by first pre-training on Kinetics and then fine-tuning on target dataset, i.e. all three splits of the UCF101 and HMDB51 datasets. For UCF101 and HMDB51, we report the average accuracy over all three splits. Our experiments are best with STCnet and 3D-ResNet/Next configuration which is of depth 101.

Table~\ref{table:kinetics_results} presents Kinetics dataset results for DynamoNet compared with the other 3D ConvNets who have provided results on the dataset. The DynamoNet~(STCnet101) with 64 frames input depth outperforms STC-ResNext101~\cite{stcnn} which has the input size of 32 frames, and also I3D~\cite{i3d} with 64 frames input.

\begin{table}[htb]

{\small
\tabcolsep=0.2cm
\begin{center}
\begin{tabular}{ l |   c |  c   }
\hline
\textbf{Method} &  \textbf{UCF101} & \textbf{HMDB51}\\
\hline
DT+MVSM~\cite{dtmvsv}       & 83.5 & 55.9 \\ \hline
iDT+FV~\cite{idt}         & 85.9 & 57.2 \\ \hline
C3D~\cite{c3d}            & 82.3 & 56.8 \\ \hline
C3D+iDT~\cite{c3d}            & 90.4 & $-$  \\ \hline
LTC+iDT~\cite{c3d}            & 92.4 & 67.2 \\ \hline
Conv Fusion~\cite{pooling}          & 82.6 & 56.8 \\ \hline
Two Stream~\cite{twostream}   & 88.6 & $-$  \\ \hline
TDD+FV~\cite{tdd}         & 90.3 &  63.2  \\ \hline
RGB+Flow-TSN~\cite{tsn}           & 94.0 & 68.5 \\ \hline
ST-ResNet~\cite{spatialRes} & 93.5 & 66.4 \\ \hline
TSN~\cite{tsn}  & 94.2 & 69.5 \\ \hline
RGB-I3D \cite{i3d}  & 95.6 & {74.8} \\ \hline

Inception3D \cite{stcnn}  & 87.2 & 56.9 \\ \hline

3D ResNet 101 (16 frames) \cite{3DResHara}  & 88.9 & 61.7 \\ \hline
3D ResNext 101 (16 frames) \cite{3DResHara} & 90.7 & 63.8 \\ \hline
STC-ResNext 101 (16 frames) \cite{stcnn}  & {92.3}    &  {65.4} \\ \hline
{STC-ResNext 101 (64 frames)} \cite{stcnn}  & {96.5}    &  {74.9} \\ \hline

\hline\hline
 
\textbf{DynamoNet~(ResNext) (16 frames)}          & \textbf{91.6}   & \textbf{66.2}\\ \hline
\textbf{DynamoNet~(ResNext) (32 frames)}          & \textbf{93.1}   & \textbf{68.5}\\ \hline
\textbf{DynamoNet~(STCnet) (32 frames)}         & \textbf{96.6}   & \textbf{74.9}\\ \hline
\textbf{DynamoNet~(STCnet) (64 frames)}         & \textbf{97.8}   & \textbf{76.8}\\ \hline

\end{tabular}
\end{center}}
\caption{Accuracy~(\%) performance comparison of \textbf{DynamoNet} with the state-of-the-art methods over all three splits of UCF101 and
HMDB51. For a fair comparison, in this table we report the performance of methods which utilize only RGB frames as input.}
\label{table:state_comparison_UCFHMDB}
\end{table}

%

In Table~\ref{table:state_comparison_UCFHMDB}, we compare the performance of DynamoNet with current state-of-the-art methods on UCF101/HMDB51. Our DynamoNet (with STCnet model) outperforms STCnet~\cite{stcnn}, 3D-ResNet~\cite{res3d}, RGB-I3D\cite{i3d} and C3D~\cite{c3d} on both UCF101 and HMDB51 and achieves 97.8\% and 76.8\% accuracy respectively. As shown in Table~\ref{table:state_comparison_UCFHMDB}, DynamoNet performs better than STC-ResNext101 by almost 2\% on UCF101. Note that most of current methods~\cite{i3d,tsn} utilize optical-flow maps in addition to RGB frames,
such as I3D which obtains a performance of 98\% on UCF101 and 80\% on HMDB51, also utilize flow information. 
Since our DynamoNet is providing motion representation for action classification in an end-to-end fashion, we show that a nice performance can also be obtained even without incorporating flow information.

Despite of not using optical-flow information, our results show how DynamoNet can exploit spatio-temporal appearance and motion information together with 3D-Conv structure, in addition to dynamic motion representation learning. Our work encourages similar approaches to exploit motion cues for action and activity classification in a more efficient manner thus leading to improve both accuracy and computation performance.

\section{Conclusion} \label{sec:conc}
The capability of the current video understanding architectures to effectively learn and exploit motion representation is a key issue in the field of action classification. In this work, we propose to learn an action classification driven motion representation in videos using dynamic motion filters by predicting the future frames. Furthermore, we show that the learned motion representation is effective for action classification. 
We demonstrate the effectiveness of our proposed method on three challenging action recognition benchmark datasets: UCF101, HMDB51 and Kinetics. In addition to yielding a better performance than the state-of-the-art methods, our dynamic motion representations are robust and compact - which retains a global motion representation in a more expressive way. 

Even though, in this paper we have focused on action classification only, we believe the our motion information can be added as a complementary cue for other tasks, like video understanding, video retrieval and more. Since our motion filter learning is self-supervised, we believe abundantly available unlabeled videos are an effective resource to acquire knowledge and to learn a effective feature representation. In future, we would like to explore two-stream network paradigm for making a more efficient pipeline of frame prediction and action classification.

\noindent
\textbf{Acknowledgements:} This work was supported by DBOF PhD scholarship, KU Leuven:CAMETRON project, and KIT:DFG-PLUMCOT project. 

{\small
\bibliographystyle{ieee}
\bibliography{egbib}
}

\end{document}